%% file: recipeUnderstanding.tex
\documentclass[10pt,twocolumn,letterpaper]{article}

\usepackage{iccv}
\usepackage{times}
\usepackage{epsfig}
\usepackage{graphicx}
\usepackage{amsmath}
\usepackage{amssymb}
\usepackage{dsfont}
\usepackage{xcolor}
 \usepackage{setspace}

 \usepackage[small,it]{caption}
 \usepackage{subcaption}
\graphicspath{ {./images/} }

\DeclareMathOperator*{\argmax}{arg\,max}

\usepackage[pagebackref=true,breaklinks=true,letterpaper=true,colorlinks,bookmarks=false]{hyperref}

\iccvfinalcopy 

\def\iccvPaperID{623} 

\ificcvfinal\fi
\begin{document}

\title{Unsupervised Semantic Parsing of Video Collections}

\author{ Ozan Sener$^{1,2}$ \;\; Amir R. Zamir$^{1}$ \;\; Silvio Savarese$^{1}$ \;\; Ashutosh Saxena$^{2,3}$  \\ \\
$^1$ Department of Computer Science, Stanford University\\
$^2$ Department of Computer Science, Cornell University\\
$^3$ Brain of Things Inc.\\
{\tt\small \{ozansener,zamir,ssilvio,asaxena\}@cs.stanford.edu}
}
\maketitle

\begin{spacing}{0.98}
\begin{abstract}
\vspace{-2mm}
Human communication typically has an underlying structure. This is reflected in the fact that in many user generated videos, a starting point, ending, and certain objective steps between these two can be identified. In this paper, we propose a method for parsing a video into such semantic steps in an unsupervised way. The proposed method is capable of providing a ``semantic storyline" of the video composed of its objective steps. We accomplish this using both visual and language cues in a joint generative model. The proposed method can also provide a textual description for each of the identified semantic steps. We evaluate this method on a large number of complex YouTube videos and show results of unprecedented quality for this intricate and impactful problem.
\end{abstract}
\vspace{-5mm}
\input{intro}
\input{related}
\input{overview}

\input{method-features}
\input{method-learning}
\input{experiments}
\input{conclusion}
\end{spacing}

\end{document}

%% file: intro.tex

\section{Introduction}
\vskip -.05in
Human communication takes many forms, including language and vision. For instance, explaining ``how-to'' perform a certain task can be communicated via language (\eg, Do-It-Yourself books) as well as visual (\eg, instructional YouTube videos) information. Regardless of the form, such human-generated communication is generally structured and has a clear beginning, end, and a set of steps in between. Parsing such communication into its semantic steps is the key to understanding structured human activities.

Language and vision provide different, but correlating and complementary information. Challenge lies in that both video frames and language (from subtitles generated via ASR) are only a noisy, partial observation of the actions being performed. However, the complementary nature of language and vision gives the opportunity to understand the activities from these partial observations. In this paper, we present a unified model, incorporating both of the modalities, in order to parse human activities into activity steps with no form of supervision other than requiring videos to be of the same category (\eg, videos retrieved by query cooking eggs, changing tires, etc.).

%




\begin{figure}[h!]
  \includegraphics[width=0.48\textwidth]{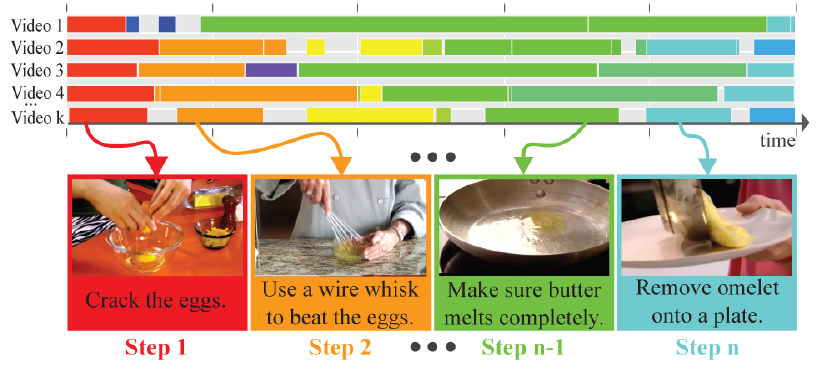}
  \vskip -.1in
  \caption{Given a large video collection (frames and subtitles) of an structured category (\eg, How to cook an omelette?), we discover activity steps (\eg, crack the eggs). We also parse the videos based on the discovered steps.}
  \label{teaser}
  \vspace{-4mm}
\end{figure}

The key idea in our approach is the observation that the large collection of videos, pertaining to the same activity class, typically include only a few objective activity steps, and the variability is the result of exponentially many ways of generating videos from activity steps through subset selection and time ordering. We study this construction based on the large-scale information available in YouTube in the form of instructional videos  (\eg, ``Making pancake'', ``How to tie a bow tie''). We adopt Instructional videos since they have many desirable properties like the volume of the information (\eg, YouTube has 281.000 videos for \emph{"How to tie a bow tie"}) and a well defined notion of activity step.  However, the proposed parsing method is applicable to any type of structured videos as long as they are composed of a set of objective steps.

The output of our method can be seen as the ``semantic storyline'' of a rather long and complex video collection (see Fig.~\ref{teaser}). This storyline provides what particular steps are taking place in the video collection, when they are occurring, and what their meaning is (\emph{what-when-how}). This method also puts videos performing the same overall task in common ground and capture their high-level relations.

In the proposed approach, given a collection of videos, we first generate a set of language and visual atoms. These atoms are the result of relating object proposals from each frame as well as detecting the frequent words from subtitles. We then employ a generative \emph{beta process mixture model}, which identifies the activity steps shared among the videos of the same category based on a representation using learned atoms. The discovered steps are found to be highly correlating with semantic steps since the semantics are the strongest common structure among all of the videos of one category. In our method, we use neither any spatial or temporal label on actions/steps nor any labels on object categories. We later learn a Markov language model to provide a textual description for each of the activity steps based on the language atoms it frequently uses.


%% file: related.tex

\section{Related Work}

Three aspects differentiate this work from the majority of existing techniques: 1) discovering semantic steps from a video category, 2) being unsupervised, 3) adopting a multi-modal joint vision-language model for video parsing. A thorough review of the related literature is provided below.

\noindent\textbf{Video Summarization:} Summarizing an input video as a sequence of key frames (static) or video clips (dynamic) is useful for both multimedia search interfaces and retrieval purposes. Early works in the area are summarized in~\cite{vidAbstraction} and mostly focus on \emph{choosing keyframes} for visualization.

Summarizing videos is particularly important for long sequences like ego-centric videos and news reports~\cite{lee2012discovering, lu2013story,rui2000automatically}; however, these methods mostly rely on characteristics of the application and do not generalize.

Summarization is also applied to the large image collections by recovering the temporal ordering and visual similarity of images \cite{storyGraph}, and by Gupta et al.~\cite{gupta2009understanding} to videos in a supervised framework using annotations of actions. These collections are also used to choose important scenes for key-frame selection \cite{khosla2013large} and further extended to video clip selection \cite{kim2014joint,potapov2014category}. Unlike all of these methods which  focus on forming a set of key frames/clips for a compact summary (which is not necessarily semantically meaningful), we provide a fresh approach to video summarization by performing it through semantic parsing on vision and language. However, regardless of this dissimilarity, we experimentally compare our method against them.

\noindent\textbf{Modeling Visual and Language Information:}
Learning the relationship between the visual and language data is a crucial problem due to its immense applications. Early methods \cite{matching} in this area focus on learning a common multi-modal space in order to jointly represent language and vision. They are further extended to learning higher level relations between object segments and words \cite{connecting}. Similarly, 
Zitnick et al.~\cite{zitnick2013learning,zitnick2013bringing} used abstracted clip-arts to understand spatial relations of objects and their language correspondences. Kong et al.~\cite{kong2014you} and 
Fidler et al.~\cite{fidler2013sentence} both accomplished the task of learning spatial reasoning using the image captions. Relations extracted from image-caption pairs, are further used to help semantic parsing \cite{yu2013grounded} and activity recognition \cite{motwani2012improving}. Recent works also focus on automatic generation of image captions with underlying ideas ranging from finding similar images and transferring their captions \cite{ordonez2011im2text} to learning language models conditioned on the image features \cite{kiros2014multimodal,socher2014grounded,farhadi2010every}; their employed approach to learning language models is typically either based on graphical models \cite{farhadi2010every} or neural networks \cite{socher2014grounded,kiros2014multimodal,deepAlignment}.

All aforementioned methods use supervised labels either as strong image-word pairs or weak image-caption pairs, while our method is fully unsupervised.

\noindent\textbf{Activity/Event Recognition:}
The literature of activity recognition is broad. The closest techniques to ours are either supervised or focus on detecting a particular (and often short) action in a weakly/unsupervised manner. Also, a large body of action recognition methods are intended for trimmed videos clips or remain limited to detecting very short actions~\cite{kuehne2011hmdb, UCF101, niebles10_eccv, laptev08_cvpr, efros03_iccv, ryoo09_iccv}. Even though some recent works attempted action recognition in untrimmed videos~\cite{THUMOS14, oneata2014lear, jainuniversity}, they are mostly fully supervised.

Additionally, several method for localizing instances of actions in rather longer video sequences have been developed~\cite{duchenne09_iccv, hoai11_cvpr, laptev07_iccv, bojanowski14_eccv, pirsiavash14_cvpr}. Our work is different from those in terms of being multimodal, unsupervised, applicable to a video collection, and not limited to identifying predefined actions or the ones with short temporal spans.
Also, the previous works on finding action primitives such as~\cite{niebles10_eccv, yao10b_cvpr, jain13_cvpr,lan14_eccv, lan14_vs} are primarily limited to discovering atomic sub-actions, and therefore, fail to identify complex and high-level parts of a long video.

Recently, event recounting has attracted much interest and intends to identify the evidential segments for which a video belongs to a certain class~\cite{sun2014discover,das2013thousand,barbu2012video}. Event recounting is a relatively new topic and the existing methods mostly employ a supervised approach. Also, their end goal is to identify what parts of a video are highly related to an event, and not parsing the video into semantic steps.


\noindent\textbf{Recipe Understanding:}
Following the interest in community generated recipes in the web, there have been many attempts to automatically process recipes. Recent methods on natural language processing \cite{cookingSemantics,logicRecipe} focus on semantic parsing of language recipes in order to extract actions and the objects in the form of predicates. Tenorth et al.~\cite{logicRecipe} further process the predicates in order to form a complete logic plan. The aforementioned approaches focus only on the language modality and they are not applicable to the videos. The recent advances \cite{beetz,cookie} in robotics use the parsed recipe in order to perform cooking tasks. They use supervised object detectors and report a successful autonomous experiment. In addition to the language based approaches, Malmaud et al.~\cite{alignment} consider both language and vision modalities and propose a method to align an input video to a recipe. However, this method can not extract the steps automatically and requires a ground truth recipe to align. On the contrary, our method uses both visual and language modalities and extracts the actions while autonomously discovering the steps. Also, \cite{photoshop} generates multi-modal recipes from expert demonstrations . However, it is developed only for the domain of ``teaching user interfaces" and are not applicable to videos.

%% file: overview.tex

\vspace{-1mm}
\section{Overview}
\label{sec:overview}
\vspace{-1mm}


Given a large video-collection, our algorithm starts with learning a set of visual and language atoms which are further used for representing multimodal information (Section~\ref{atoms}). These atoms are designed to be more likely to correspond to the mid-level semantic concepts like actions and objects. In order to learn visual atoms, we generate object proposals and cluster them into mid-level atoms. Whereas, for the language atoms we simply use the salient and frequent words in the subtitles. After learning the atoms, we represent the multi-modal information in each frame based on the occurrence statistics of the atoms (Section~\ref{atoms}); Given the sequence of multi-modal frame representations, we discover a set of clusters occurring over multiple videos using a non-parametric Bayesian method (Section~\ref{learning}). We expect these clusters to correspond to the activity steps which construct the high level activities. Our empirical results confirms this as the resulting clusters significantly correlates with the activity steps.



%% file: method-features.tex

\vspace{-1mm}
\section{Forming the Multi-Modal Representation}
\label{atoms}
\vspace{-1mm}

Finding the set of activity steps over large collection of videos having large visual varieties requires us to represent the semantic information in addition to the low-level visual cues. Hence, we find our language and visual atoms by using mid-level cues like object proposals and frequent words.

\begin{figure}[h]
  \includegraphics[width=0.5\textwidth]{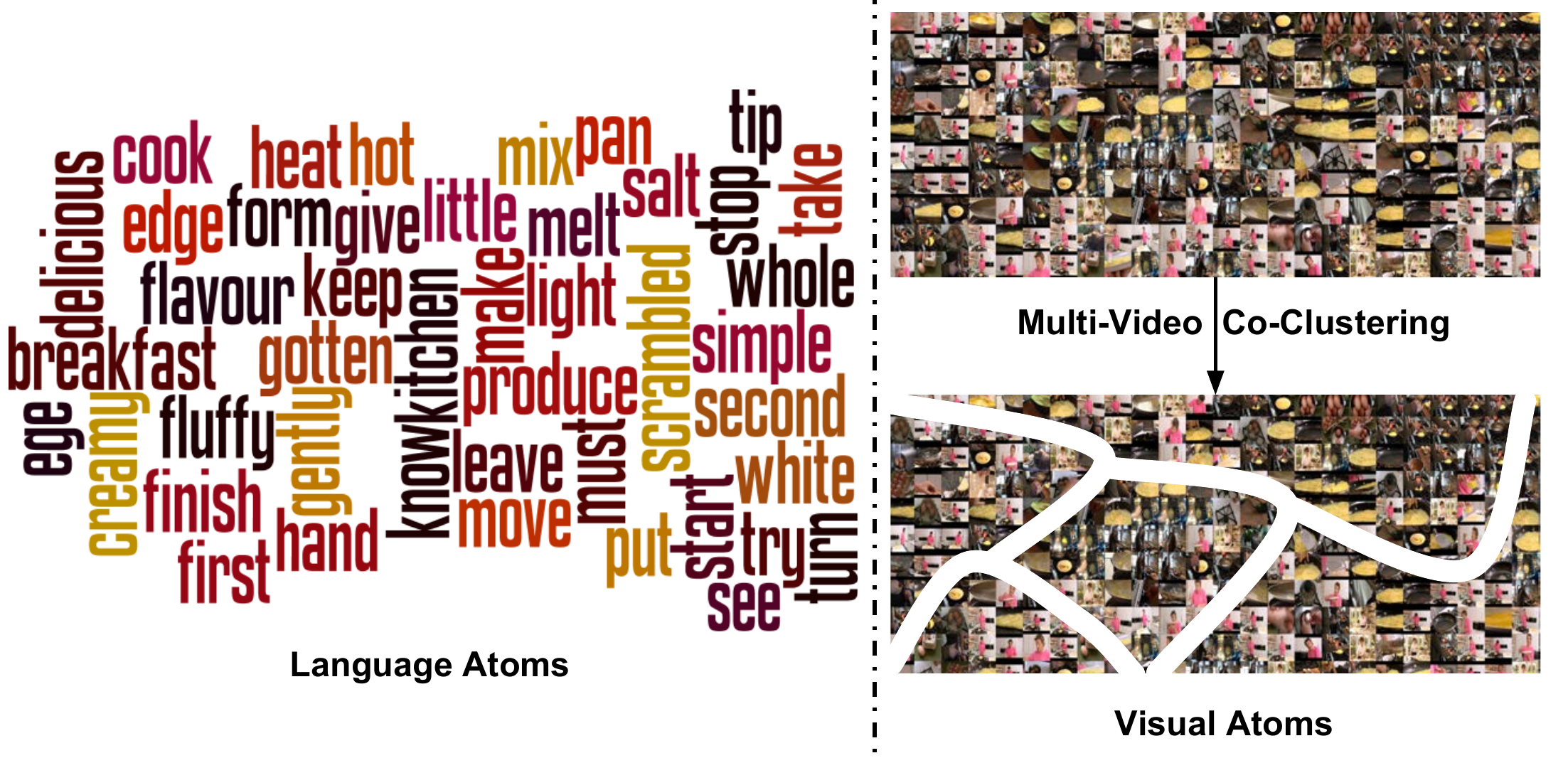}
  \vspace{-8mm}
  \caption{We learn language and visual atoms to represent multi-modal information. Language atoms are frequent words and visual atoms are the clusters of object proposals.}
  \vspace{-3mm}
\label{fig:overview}
\end{figure}

\noindent\textbf{Learning Visual Atoms:} In order to learn visual atoms, we create a large collection of  proposals by independently generating object proposals from each frame of each video. These proposals are generated using the Constrained Parametric Min-Cut (CPMC) \cite{cpmc} algorithm based on both appearance and motion cues. We note the $k^{th}$ proposal of $t^{th}$ frame of $i^{th}$ video as $r^{(i),k}_t$. Moreover, we drop the video index $(i)$ if it is clearly implied in the context.

In order to group this object proposals into mid-level visual atoms, we follow a clustering approach. Although any graph clustering approach (\eg, Keysegments \cite{keysegments}) can be applied for this, the joint processing of a large video collection requires handling \emph{large visual variability} among multiple videos. We propose a new method to jointly cluster object proposals over multiple videos in Section~\ref{jointProp}. Each cluster of object proposals correspond to a visual atom.

\noindent\textbf{Learning Language Atoms:}
We define the language atoms as the salient words which occur more often than their ordinary rates based on the \emph{tf-idf} measure. The \emph{document} is defined as the concatenation of all subtitles of all frames of all videos in the collection. Then, we follow the classical tf-idf measure and use it as $tfidf(w,D)=f_{w,D} \times \log \left( 1+ \frac{N}{n_{w}}\right)$ where $w$ is the word we are computing the tf-idf score for, $f_{w,D}$ is the frequency of the word in the \emph{document} $D$, $N$ is the total number of video collections we are processing, and $n_{w}$ is the number of video collections whose subtitle include the word $w$.

We sort words with their ``tf-idf" values and choose the top $K$ words as language atoms (\emph{$K=100$ in our experiments}). As an example, we show the language atoms learned for the category \emph{making scrambled egg} in Figure~\ref{fig:overview}. 

\begin{figure}[h!]
  \includegraphics[width=0.46\textwidth]{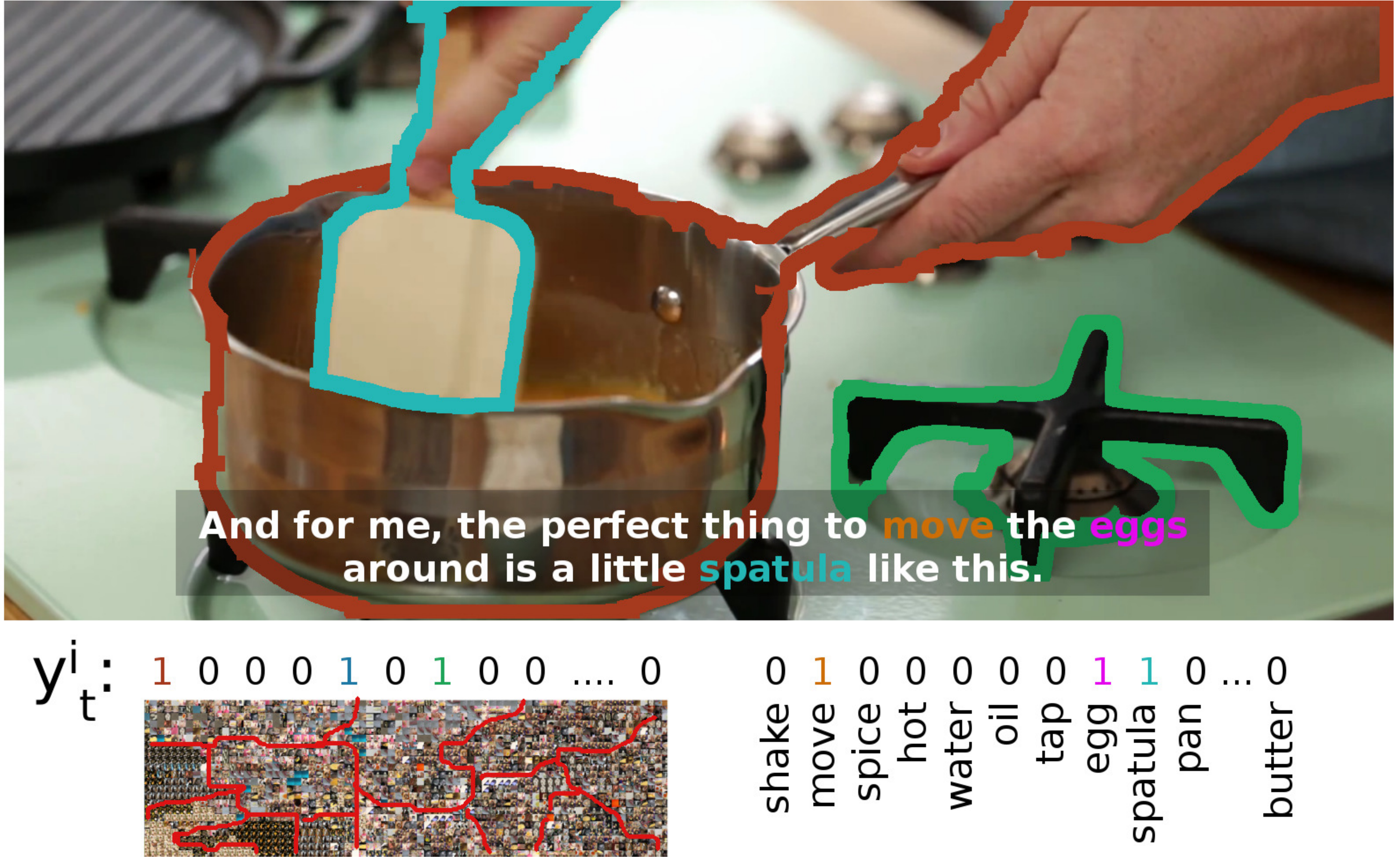}
  \caption{\textbf{Representation for a sample frame.} Three of the object proposals of sample frame are in the visual atoms and three of the words are in the language atoms.}
  \label{visFrame}
  \vspace{-4mm}
\end{figure}

\noindent\textbf{Representing Frames with Atoms:}
After learning the visual and language atoms, we represent each frame via the occurrence of atoms (binary histogram). Formally, the representation of the $t^{th}$ frame of the $i^{th}$ video is denoted as $\mathbf{y^{(i)}_t}$ and computed as $\mathbf{y^{(i)}_t}=[\mathbf{y^{(i),l}_t},\mathbf{y^{(i),v}_t}]$ such that $k^{th}$ entry of the $\mathbf{y^{(i),l}_t}$ is $1$ if the subtitle of the frame has the $k^{th}$ language atom and $0$ otherwise. $\mathbf{y^{(i),v}_t}$ is also a binary vector similarly defined over visual atoms. We visualize the representation of a sample frame in the Figure~\ref{visFrame}.

\vspace{-1mm}
\section{Joint Proposal Clustering over Videos}
\label{jointProp}
\vspace{-1mm}

Given a set of object proposals generated from \emph{multiple videos}, simply combining them into a single collection and clustering them into atoms is not desirable for two reasons: (1) semantic concepts have large visual differences among different videos and accurately clustering them into a single atom is hard, (2) atoms should contain object proposals from multiple videos in order to semantically relate the videos. In order to satisfy these requirements, we propose a joint extension to spectral clustering. Note that the purpose of this clustering is generating atoms where each clusters represents an atom.

\begin{figure}[ht]
  \includegraphics[width=0.48\textwidth]{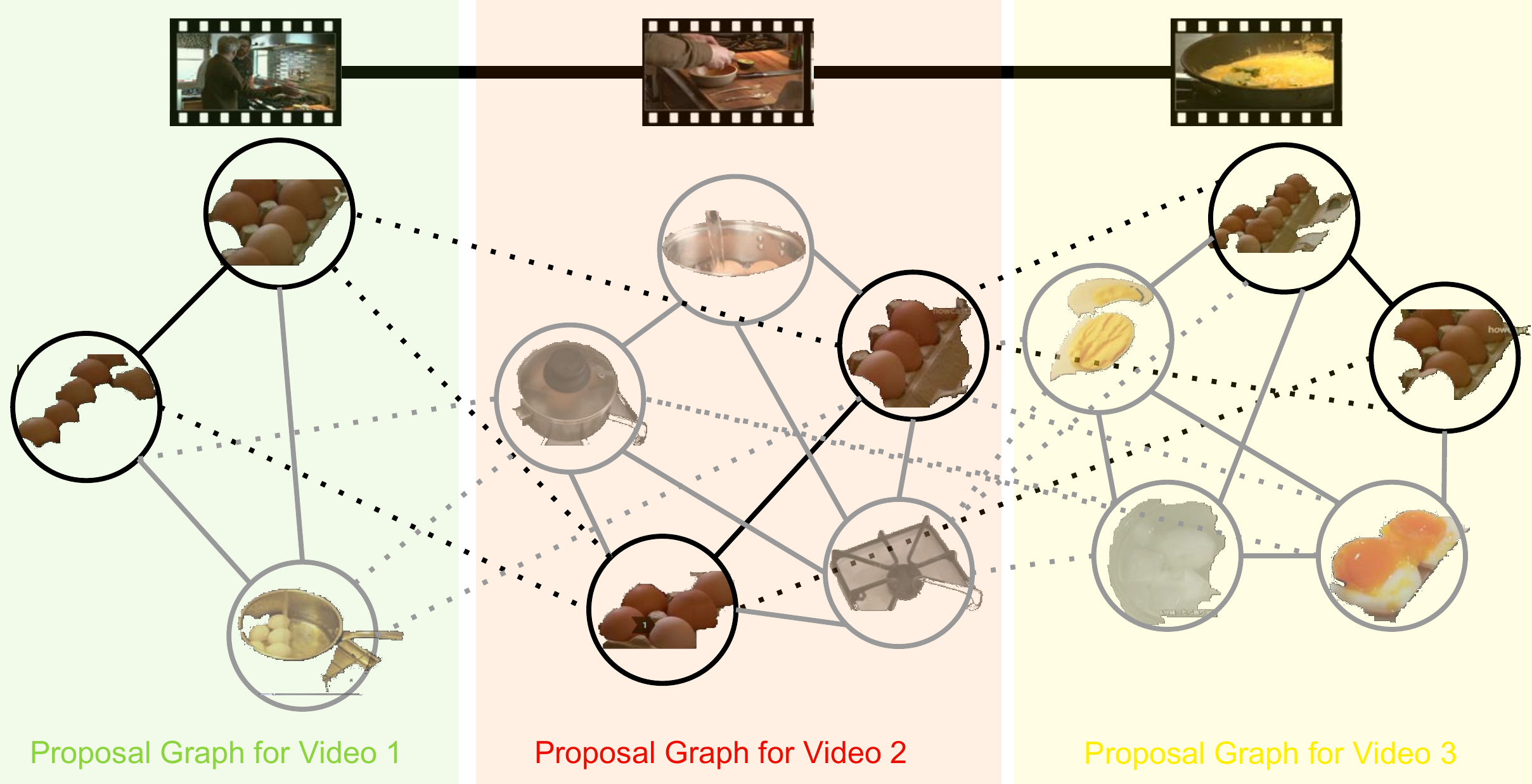}
  \vskip -2mm
\caption{\textbf{Joint proposal clustering.} Each object proposal is linked to its two NNs from the video it belongs and two NNs from the videos it is neighbour of. Dashed and solid lines denote the intra-video and inter-video edges, respectively. Black nodes are the proposals selected as part of the cluster and the gray ones are not selected. Similarly, the black and gray edges denote selected and not-selected, respectively.}
  \label{hierProposal}
    \vskip -1mm
\end{figure}

\noindent\textbf{Basic Graph Clustering:} Consider the set of object proposals extracted from a single video $\{r^k_t\}$, and a pairwise similarity metric $d(\cdot,\cdot)$ for them. Single cluster graph partitioning (SCGP)\cite{scgp} approach finds the dominant cluster which maximizes the intra-cluster similarity:
\begin{equation}
  \argmax_{x^k_t} \frac{\sum_{(k_1,t_1),(k_2,t_2) \in K \times T} x^{k_1}_{t_1} x^{k_2}_{t_2} d(r^{k_1}_{t_1},r^{k_2}_{t_2})}{\sum_{(k,t) \in K \times T} x^{k}_t},
  \label{nonvec}
\end{equation}
where $x^{k}_t$ is a binary variable which is $1$ if $r^{k}_t$ is included in the cluster, $T$ is the number of frames and $K$ is the number of clusters per frame. Adopting the vector form of the indicator variables as $\mathbf{x_{tK+k}}=x^{k}_{t}$ and the pairwise distance matrix as $\mathbf{A}_{t_1K+k_1,t_2K+k_2}=d(r^{k_1}_{t_1},r^{k_2}_{t_2})$, equation (\ref{nonvec}) can be compactly written as
$\argmax_{\mathbf{x}} \frac{\mathbf{x^T}A\mathbf{x}}{\mathbf{x^T}\mathbf{x}}$.
This can be solved by finding the dominant eigenvector of $\mathbf{x}$ after relaxing $x^{k}_t$ to $[0,1]$ \cite{scgp,scgp_eigen}. Upon finding the cluster, the members of the selected cluster are removed from the collection and the same algorithm is applied to find remaining clusters.

\noindent\textbf{Joint Clustering:} Our extension of the SCGP into multiple videos is based on the assumption that the key objects occur in most of the videos. Hence, we re-formulate the problem by enforcing the homogeneity of the cluster over all videos.

We first create a kNN graph of the videos based on the distance between their textual descriptions. We use the $\chi^2$ distance of the bag-of-words computed from the video description. We also create the kNN graph of object proposals in each video based on the pretrained "fc7" features of AlexNet~\cite{alexnet}. This hierarchical graph structure is visualized in Figure~\ref{hierProposal} for three videos samples. After creating this graph, we impose both ``inter-video" and ``intra-video" similarity among the object proposals of each cluster. Main rationale behind this construction is having a separate notion of distance for inter-video and intra-video relations since the visual similarity decreases drastically for inter-video ones.

Given the intra-video distance matrices $\mathbf{A^{(i)}}$, the binary indicator vectors $\mathbf{x^{(i)}}$, and the inter-video distance matrices as $\mathbf{A^{(i,j)}}$, we define our optimization problem as:
\begin{equation}
\argmax \sum_{i \in N} \frac{\mathbf{x^{(i)^T}}\mathbf{A^{(i)}}\mathbf{x^{(i)}}}{\mathbf{x^{(i)^T}}\mathbf{x^{(i)}}} +
\sum_{i \in N} \sum_{j \in \mathcal{N}(i)} \frac{\mathbf{x^{(i)^T}}\mathbf{A^{(i,j)}}\mathbf{x^{(j)}}} {\mathbf{x^{(i)^T}}\mathds{1}\mathds{1}^T\mathbf{x^{(j)}}},
\end{equation}
where $\mathcal{N}(i)$ is the neighbours of the video $i$ in the kNN graph, $\mathds{1}$ is vector of ones and $N$ is the number of videos.

Although we can not use the efficient eigen-decomposition approach from \cite{scgp,scgp_eigen} as a result of the modification, we can use Stochastic Gradient Descent as the cost function is quasi-convex when relaxed. We use the SGD with the following analytic gradient function:
\begin{equation}
  \nabla_{\mathbf{x^{(i)}}} = \frac{2\mathbf{A^{(i)}} \mathbf{x^{(i)}} -2\mathbf{x^{(i)}} r^{(i)}}
  {\mathbf{{x^{(i)}}^T}\mathbf{x^{(i)}}}
+ \sum_{i \in N} \frac{\mathbf{A^{i,j}}\mathbf{x^{j}} - \mathbf{{x^{(j)}}^T} \mathds{1} r^{(i,j)}}{\mathbf{{x^{(i)}}^T} \mathds{1} \mathds{1}^T \mathbf{x^{(j)}} },
\end{equation}
where $r^{(i)}=\frac{\mathbf{x^{(i)^T}}\mathbf{A^{(i)}}\mathbf{x^{(i)}}}{\mathbf{x^{(i)^T}}\mathbf{x^{(i)}}}$ and $r^{(i,j)}=\frac{\mathbf{x^{(i)^T}}\mathbf{A^{(i,j)}}\mathbf{x^{(j)}}} {\mathbf{x^{(i)^T}}\mathds{1}\mathds{1}^T\mathbf{x^{(j)}}}$

We iteratively use the method to find clusters, and stop after the $K=20$ clusters are found as the remaining object proposals were deemed not relevant to the activity. Each cluster corresponds to a visual atom for our application.

\begin{figure}[ht]
  \begin{subfigure}[b]{0.23\textwidth}
\includegraphics[width=\textwidth]{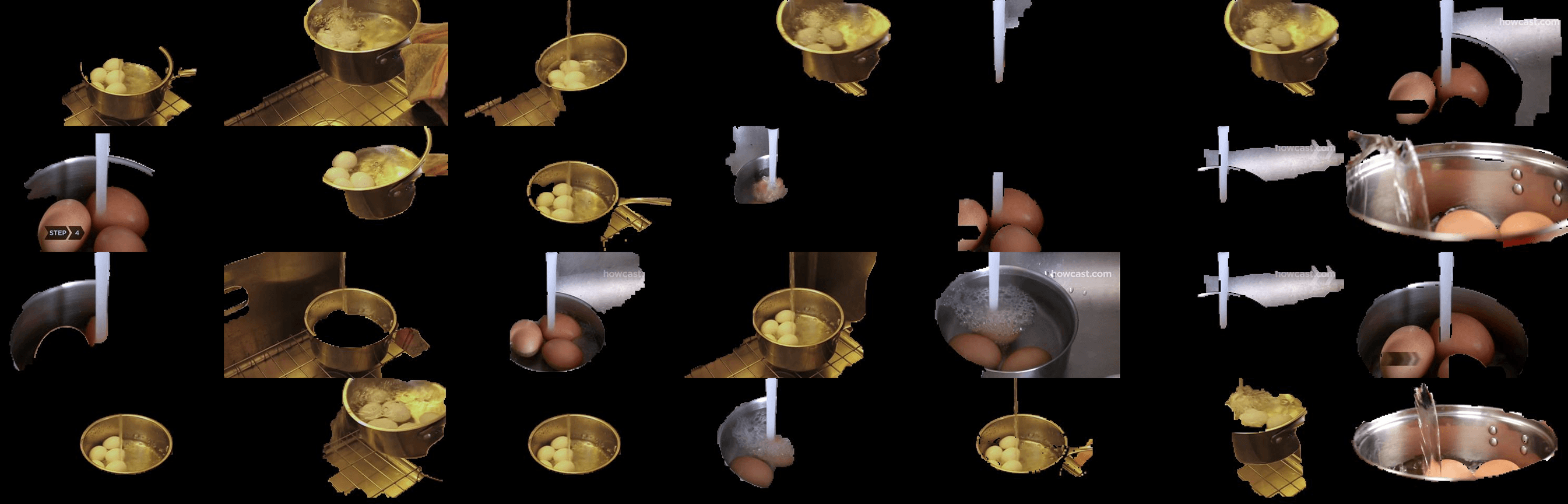}
\end{subfigure}
~
\begin{subfigure}[b]{0.23\textwidth}
\includegraphics[width=\textwidth]{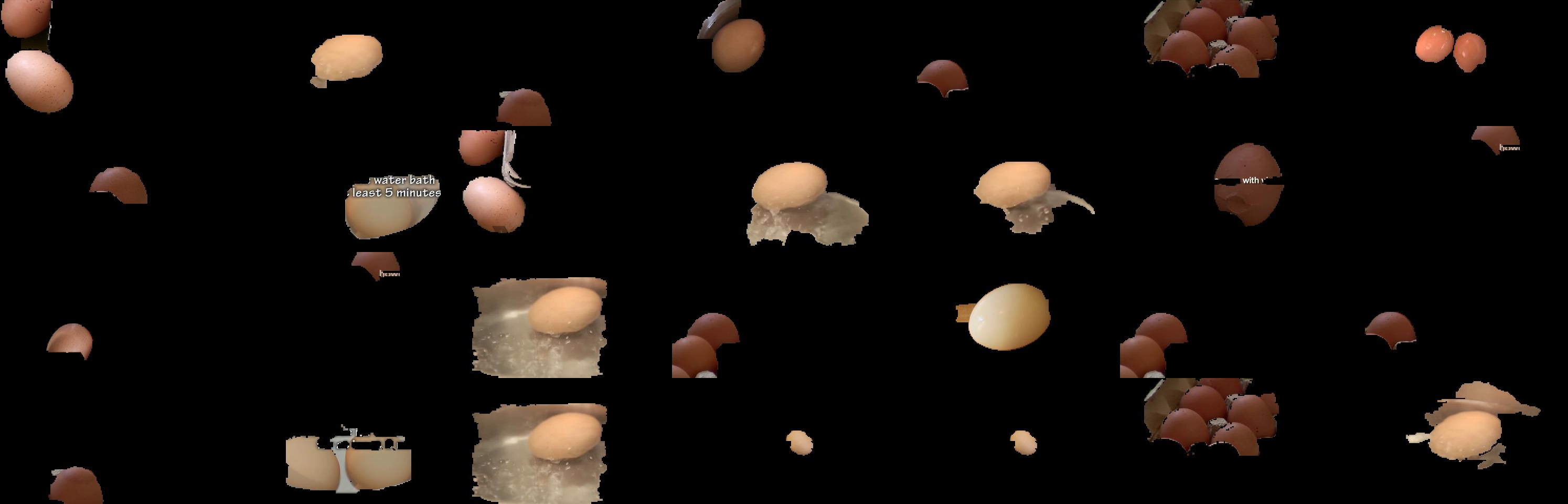}
\end{subfigure}


\begin{subfigure}[b]{0.23\textwidth}
\includegraphics[width=\textwidth]{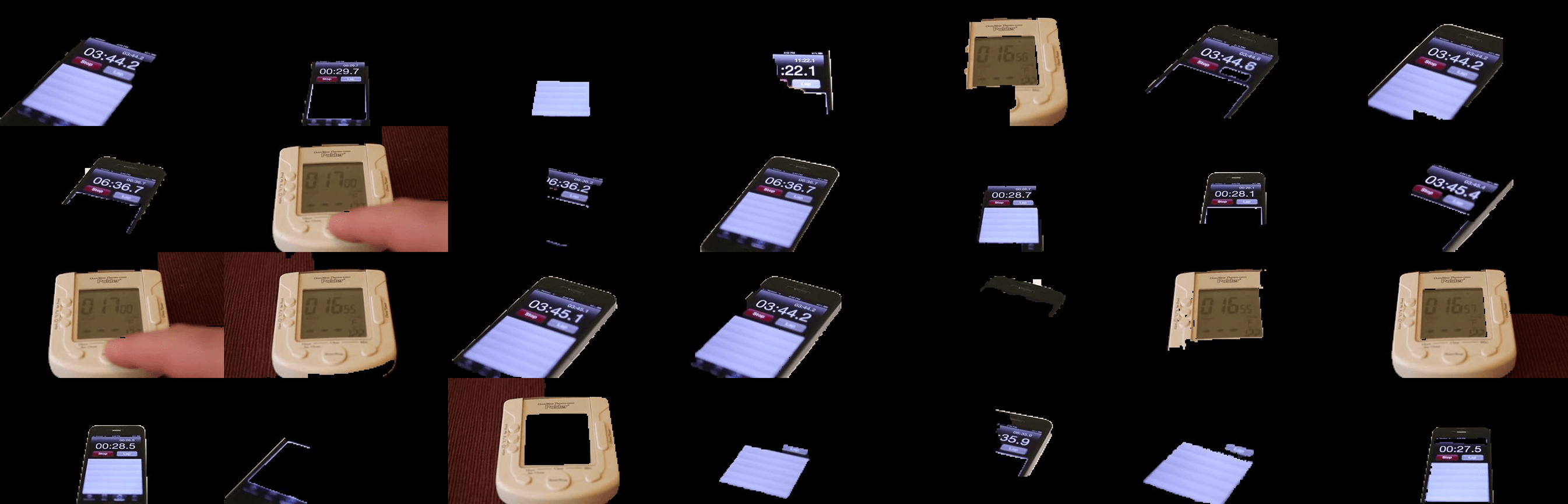}
\end{subfigure}
~
\begin{subfigure}[b]{0.23\textwidth}
\includegraphics[width=\textwidth]{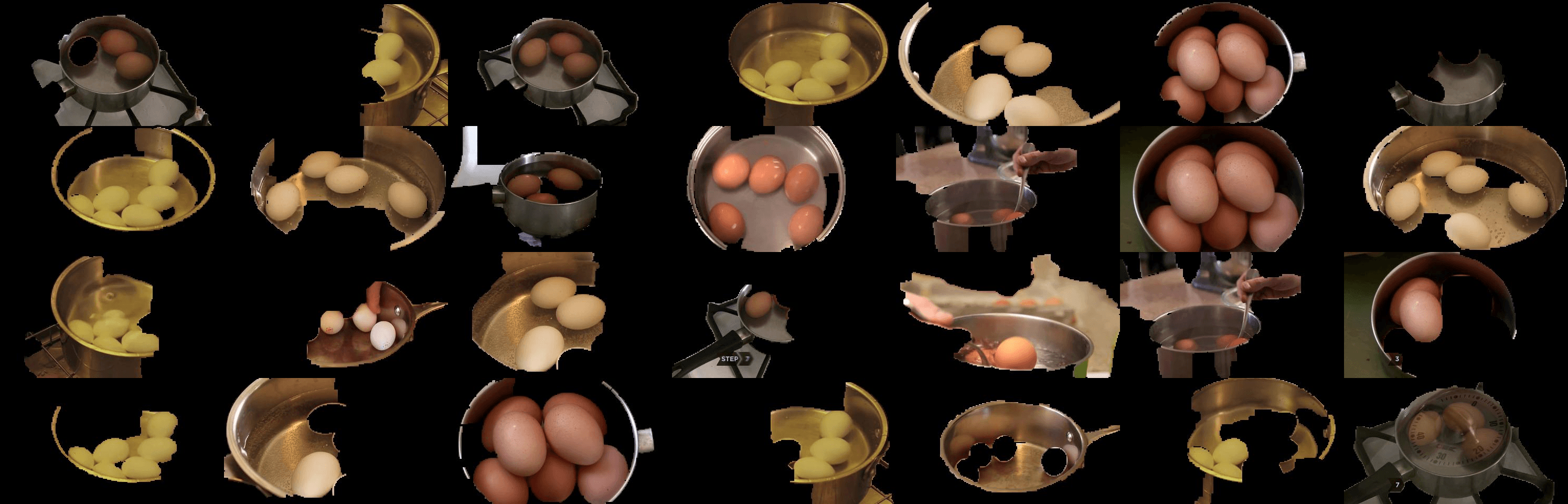}
\end{subfigure}
\vspace{-5mm}
\caption{Randomly selected images of four randomly selected clusters learned for \emph{How to hard boil an egg?}}
\label{cvis}
\vspace{-4mm}
\end{figure}

In Figure~\ref{cvis}, we visualize some of the atoms (\ie, clusters) we learned for the query \emph{How to Hard Boil an Egg?}. As apparent in the figure, the resulting atoms are highly correlated and correspond to semantic objects\&concepts regardless of their significant intra-class variability.

%% file: method-learning.tex

\subsection{Unsupervised Parsing}
\label{basics}
\label{learning}
In this section, we explain the model which we use to discover the activity steps from a video collection given the language and visual atoms. We note the extracted representation of the frame $t$ of video $i$ as $\mathbf{y^{(i)}_t}$. We model our algorithm based on activity steps and note the activity label of the $t^{th}$ frame of the $i^{th}$ video as $z^{(i)}_t$. We do not fix the the number of activities and use a non-parametric approach.


In our model, each activity step is represented over the atoms as the likelihood of including them. In other words, each activity step is a Bernoulli distribution over the visual and language atoms as $\theta_k=[\theta_k^l,\theta_k^v]$ such that $m^{th}$ entry of the $\theta_k^l$ is the likelihood of observing $m^{th}$ language atom in the frame of an activity $k$. Similarly, $m^{th}$ entry of the $\theta_k^v$ represents the likelihood of seeing $m^{th}$ visual atom. In other words, each frame's representation $\mathbf{y^{(i)}_t}$ is sampled from the distribution corresponding to its activity as \mbox{$\mathbf{y^{(i)}_t}|z^{(i)}_t=k \sim Ber(\theta_k)$}. As a prior over $\theta$, we use its conjugate distribution -- \emph{Beta distribution}.

Given the model above, we explain the generative model which links activity steps and frames in Section~\ref{bphmm}.
 
   \vskip -2mm
\subsubsection{Beta Process Hidden Markov Model}
\label{bphmm}
  \vskip -2mm
  
For the understanding of the time-series information, Fox et al.~\cite{foxBPHMM} proposed the Beta Process Hidden Markov Models (BP-HMM). In BP-HMM setting, each time-series exhibits a subset of available features. Similarly, in our setup each video exhibits a subset of activity steps.

Our model follows the construction of Fox et al.~\cite{foxBPHMM} and differs in the choice of probability distributions since \cite{foxBPHMM} considers Gaussian observations while we adopt binary observations of atoms. In our model, each video $i$ chooses a set of activity steps through an activity step vector $\mathbf{f^{(i)}}$ such that $f^{(i)}_k$ is $1$ if $i^{th}$ video has the activity step $k$, and 0 otherwise. When the activity step vectors of all videos are concatenated, it becomes an activity step matrix $\mathbf{F}$ such that $i^{th}$ row of the $\mathbf{F}$ is the activity step vector $\mathbf{f^{(i)}}$. Moreover, each activity step $k$ also has a prior probability $b_k$  and a distribution parameter $\theta_k$ which is the Bernoulli distribution as we explained in the Section~\ref{basics}.

In this setting, the activity step parameters $\theta_k$ and $b_k$ follow the \emph{beta process} as;
\vskip -2mm
\begin{equation}
  B|B_0,\gamma,\beta \sim \text{BP}(\beta,\gamma B_o), B=\sum_{k=1}^\infty b_k \delta_{\theta_k}
\end{equation}
where $B_0$ and the $b_k$ are determined by the underlying Poisson process \cite{ibp} and the feature vector is determined as independent Bernoulli draws as $f_{k}^{(i)} \sim Ber(b_k)$. After marginalizing over the $b_k$ and $\theta_k$, this distribution is shown to be equivalent to Indian Buffet Process (IBP)~\cite{ibp}. In the IBP analogy, each video is a customer and each activity step is a dish in the buffet. The first customer (video) chooses a $\text{Poisson}(\gamma)$ unique dishes (activity steps). The following customer (video) $i$ chooses previously sampled dish (activity step) $k$ with probability $\frac{m_k}{i}$,  proportional to the number of customers ($m_k$) chosen the dish $k$, and it also chooses $\text{Poisson}(\frac{\gamma}{i})$ new dishes (activity steps). Here, $\gamma$ controls the number of selected activities in each video and $\beta$ promotes the activities getting shared by videos.

The above IBP construction represents the activity step discovery part of our method. In addition, we need to model the video parsing over discovered steps; these two need to be modeled jointly. We model the each video as an Hidden Markov Model (HMM) over the selected activity steps. Each frame has the hidden state --activity step-- ($z^{(i)}_t$) and we observe the multi-modal frame representation $\mathbf{y^{(i)}_t}$. Since we model each activity step as a Bernoulli distribution, the emission probabilities follow the Bernoulli distribution as $p(\mathbf{y^{(i)}_t}|z^{(i)}_t)=Ber(\theta_{z^{(i)}_t})$.

\begin{figure}[h!]
  \includegraphics[width=0.5\textwidth]{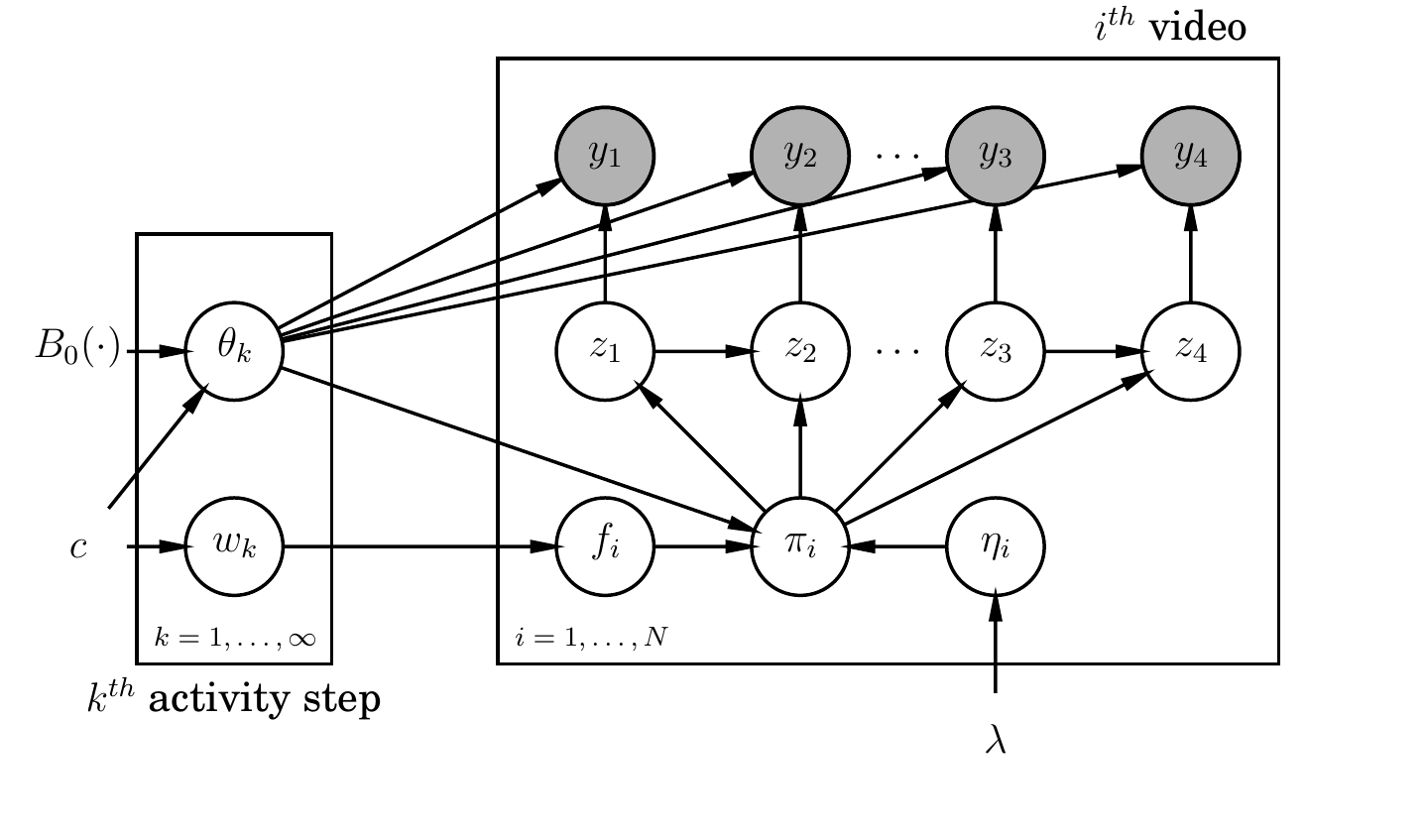}
  \vspace{-9mm}
  \caption{\textbf{Graphical model for BP-HMM:} The left plate represent the activity steps and the right plate represent the videos. (\ie the left plate is for the activity step discovery and right plate is for parsing.) \emph{See Section~\ref{bphmm} for details.}}
  \vspace{-5mm}
  \label{bphmmo}
\end{figure}

For the transition probabilities of the HMM, we do not put any constraint and simply model it as any point from a probability simplex which can be sampled by drawing a set of Gamma random variables and normalizing them \cite{foxBPHMM}. For each video $i$, a Gamma random variable is sampled for the transition between activity step $j$ and activity step $k$ if both of the activity steps are included in the video (\ie if $f^i_k$ and $f^i_j$ are both $1$). After sampling these random variables, we normalize them to make transition probabilities to sum to 1. This procedure can be represented formally as
\begin{equation}
  \eta_{j,k}^{(i)} \sim Gam(\alpha+\kappa \delta_{j,k},1), \quad \mathbf{\pi_j^{(i)}} = \frac{\mathbf{\eta^{(i)}_j} \circ \mathbf{f^{(i)}}}{\sum_k \eta^{(i)}_{j,k} f^{(i)}_k},
\end{equation}
where $\kappa$ is the persistence parameter promoting the self state transitions a.k.a. more coherent temporal boundaries, $\circ$ is the element-wise product, and $\pi^i_j$ is the transition probabilities in video $i$ from activity step $j$ to other steps. This model is also presented as a graphical model in Figure~\ref{bphmmo}.

\subsubsection{Gibbs sampling for BP-HMM}
We employ Markov Chain Monte Carlo (MCMC) method for learning and inference of the BP-HMM. We base our algorithms on the MCMC procedure proposed by Fox et al.~\cite{foxBPHMM}. Our sampling procedure composed of two samplers: (1) activity step ($\mathbf{f^{(i)}}$) sampler from the current activity step distributions $\mathbf{\theta_k}$ and multi-modal frame representations $y^{(i)}_k$, (2) and HMM parameter $\mathbf{\eta}$,$\mathbf{\pi}$,$\mathbf{\theta_k}$ sampler from the selected activities $\mathbf{f^{(i)}}$. Intuitively, we iterate over discovering activity steps given the temporal activity labels and estimating activity labels given the discovered activities. We give the details of this sampler in \cite{supp}.

%% file: experiments.tex

\section{Experiments}
In order to experiment the proposed method, we first collected a dataset (details in Section~\ref{dataset:sec}). We labelled a small part of the dataset with frame-wise activity step labels and used the resulting set as a test corpus. Neither the set of labels, nor the temporal boundaries are exposed to our algorithm since the setup is completely unsupervised. We evaluate our algorithm against the several unsupervised clustering baselines and state-of-the-art algorithms from video summarization literature which are applicable.
\subsection{Dataset}
\label{dataset:sec}
We use WikiHow~\cite{wikiHow} in order to obtain the top100 queries the internet users are interested in and choose the ones which are
directly related to the physical world. Resulting queries are;

\emph{\textbf{How to}}\footnotesize
\emph{Bake Boneless Skinless Chicken, Tie a Tie, Clean a Coffee Maker, Make Jello Shots, Cook Steak, Bake Chicken Breast, Hard Boil an Egg, Make Yogurt, Make a Milkshake, Make Beef Jerky, Make Scrambled Eggs, Broil Steak, Cook an Omelet, Make Ice Cream, Make Pancakes, Remove Gum from Clothes, Unclog a Bathtub Drain}
\normalsize

For each of the queries, we crawled YouTube and got the top 100 videos. We also downloaded the English subtitles if they exist. For the test set, we randomly choose 5 videos out of 100 per query.

\vspace{-1mm}
\subsubsection{Outlier Detection}
\label{filter}
\begin{figure}[ht]
    \includegraphics[width=0.5\textwidth]{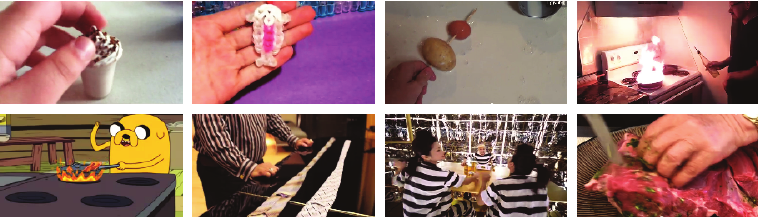}
\caption{\textbf{Sample videos which our algorithm discards as an outlier for various queries.}
A toy milkshake, a milkshake charm, a funny video about How to NOT make smoothie, a video about the danger of a fire, a cartoon video, a neck-tie video erroneously labeled as bow-tie, a song, and a lamb cooking mislabeled as chicken.}
\label{outliers}
\vspace{-3mm}
\end{figure}

\vspace{-1mm}
Since we do not have any expert intervention in our data collection, the resulting collection might have outliers, mainly due to fact that our queries are typical daily activities and there are many cartoons, funny videos, and music videos about them. Hence, we have an automatic coarse filtering stage. The key-idea behind the filtering algorithm is the fact that instructional videos have a distinguishable text descriptions when compared with outliers. Hence, we use a clustering algorithm to find the dominating cluster of instructional videos free of outliers. Given a large video collection, we use the graph, explained in Section~\ref{jointProp}, and compute the dominant video cluster by using the Single Cluster Graph Partitioning \cite{scgp} and discards the remaining videos as outlier. In Figure~\ref{outliers}, we visualize some of the discarded videos. Although our algorithm have a few percentage of false positives while detecting outliers, we always have enough number of videos (minimum 50) after the outlier detection, thanks to the large-scale dataset.

\vspace{-1mm}
\subsection{Qualitative Results}
\begin{figure*}[ht]
  \begin{subfigure}[b]{\textwidth}
    \includegraphics[width=\textwidth]{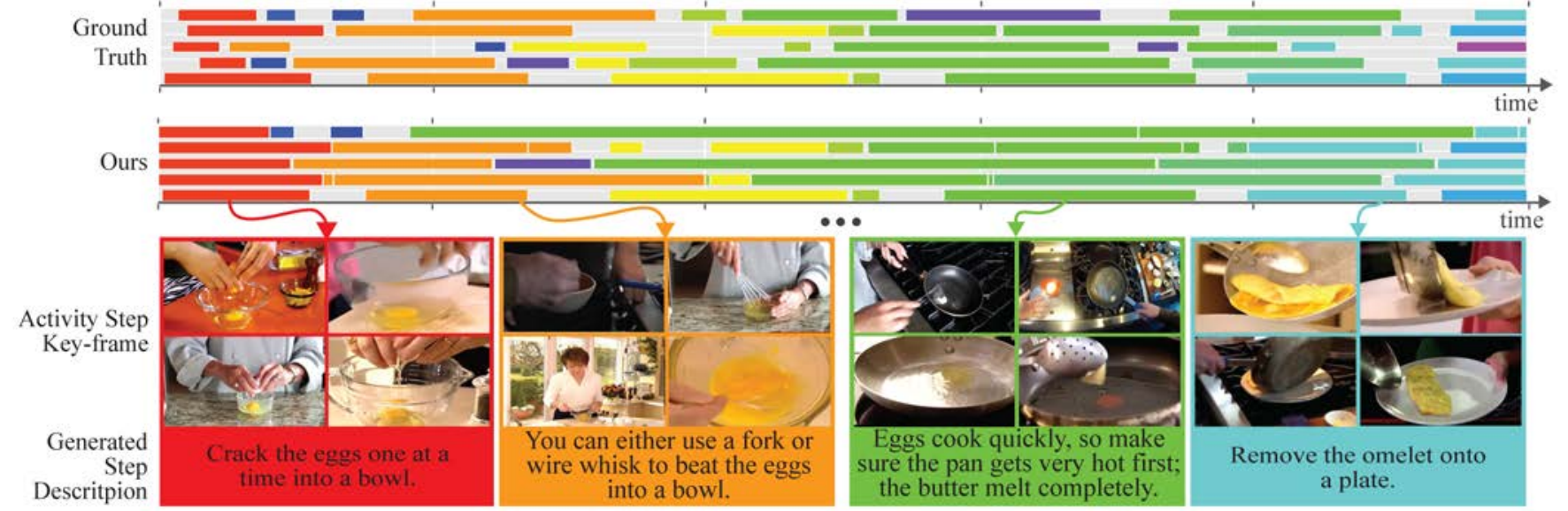}
    \vspace{-5mm}
    \caption{How to make an omelet?}
    \vspace{-1mm}
    \label{recipe:ommelette}
  \end{subfigure}

  \begin{subfigure}[b]{\textwidth}
    \includegraphics[width=\textwidth]{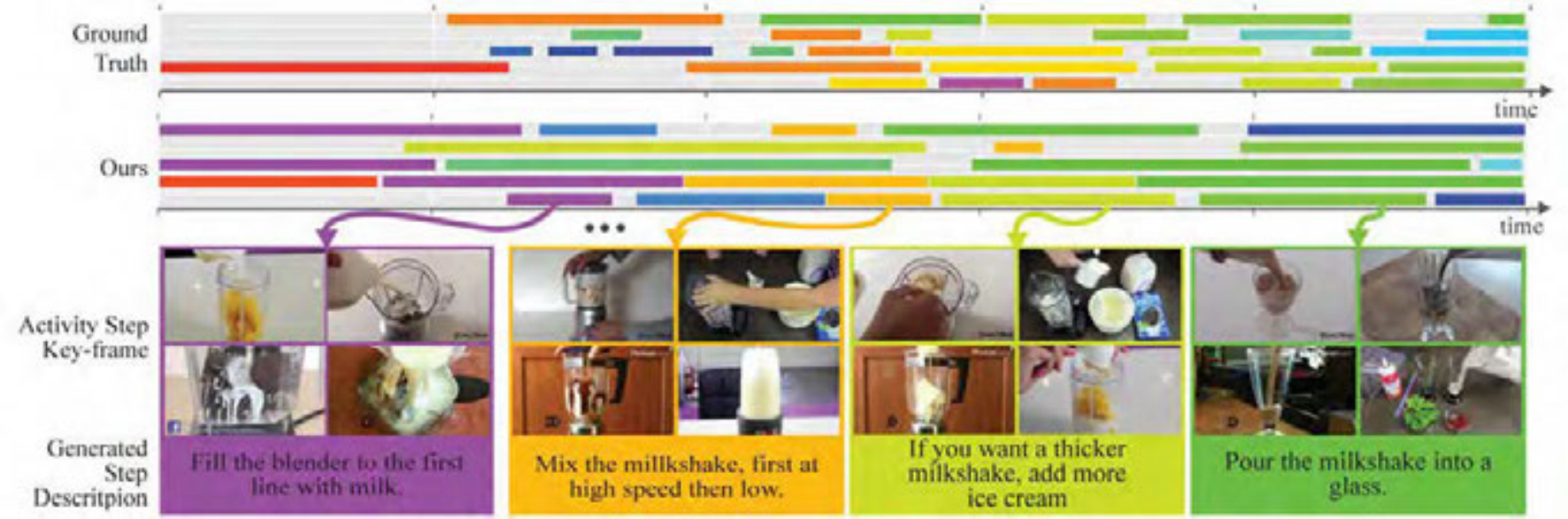}
    \caption{How to make a milkshake?}
    \vspace{-3mm}
    \label{recipe:milkshake}
  \end{subfigure}~
\caption{Temporal segmentation of the videos and ground truth segmentation. We also color code the activity steps we discovered and visualize their key-frames and the automatically generated captions. \emph{Best viewed in color.}}
\label{recipe:overall}
\vspace{-3mm}
\end{figure*}

\vspace{-1mm}
After independently running our algorithm on all categories, we discover activity steps and parse the videos according to discovered steps. We visualize some of these categories qualitatively in Figure~\ref{recipe:overall} with the temporal parsing of evaluation videos as well as the ground truth parsing.

To visualize the content of each activity step, we display key-frames from different videos. We also train a $3^{rd}$ order Markov language model~\cite{languageModel} using the subtitles and employ it to generate a caption for each activity step by sampling this model conditioned on the $\theta^l_k$. We explain the details of this process in \cite{supp}.

As shown in the Figures~\ref{recipe:ommelette} and~\ref{recipe:milkshake}, resulting steps are semantically meaningful; hence, we conclude that there is enough language context within the subtitles in order to detect activities. However, some of the activity steps occur together and our algorithm merges them into a single step as a result of promoting sparsity.

\subsection{Quantitative Results}
We compare our algorithm with the following baselines.

\noindent\textbf{Low-level features (LLF):}
In order to experiment the effect of learned atoms, we compare them low-level features. As features, we use the Fisher vector representation of Dense Trajectory like features (HOG, HOF, and MBH) \cite{kantorov2014}.

\noindent\textbf{Single modality:}
To experiment the effect of multi-modal approach, we compare with single modalities by only using the atoms of one modality.

\noindent\textbf{Hidden Markov Model (HMM):}
To experiment the effect of joint generative model, we compare our algorithm with an HMM (using the Baum-Welch \cite{rabiner} via cross-validation).

\noindent\textbf{Kernel Temporal Segmentation \cite{potapov2014category}:}
Kernel Temporal Segmentation (KTS) proposed by Potapov et al.~\cite{potapov2014category} can detect the temporal boundaries of the events/activities in the video from a time series data without any supervision. It enforces a local similarity of each resultant segment.

\begin{figure*}[t]
  \includegraphics[width=\textwidth]{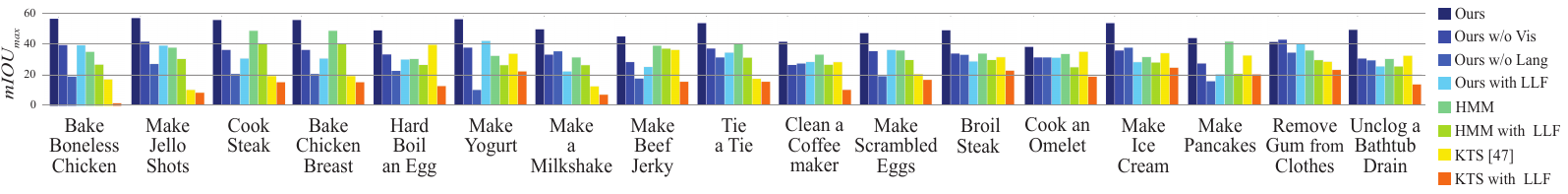}
  \vspace{-9mm}
  \caption{$IOU_{cms}$ values for all categories, for all competing algorithms.}
  \label{mIOU}
\includegraphics[width=\textwidth]{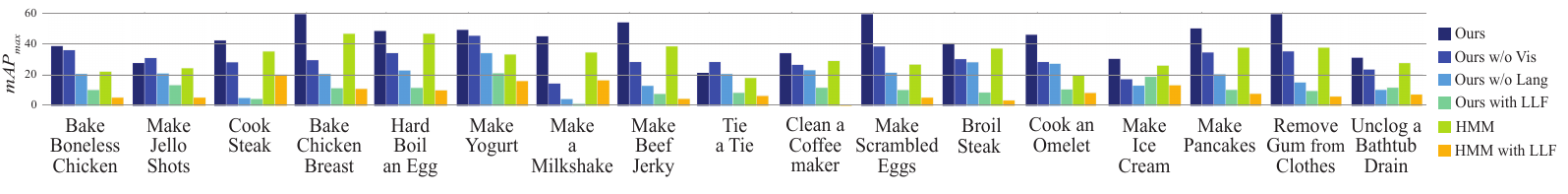}
\vspace{-9mm}
\caption{$AP_{cms}$ values for all categories, for all competing algorithms.}
\vspace{-3mm}
\label{mmAP}
\end{figure*}

Given parsing results and the ground truth, we evaluate both the quality of temporal segmentation and the activity step discovery. We base our evaluation on two widely used metrics; intersection over union ($\mathbf{IOU}$) and mean average precision($mAP$). $\mathbf{IOU}$ measures the quality of temporal segmentation and it is defined as; $\frac{1}{N}\sum_{i=1}^N \frac{\tau^\star_i \cap \tau^\prime_{i}}{\tau^\star_i \cup \tau^\prime_{i}}$ where $N$ is the number of segments, $\tau^\star_i$ is ground truth  segment and $\tau^\prime_{i}$ is the detected segment. $\mathbf{mAP}$ is defined per activity step and can be computed based on a precision-recall curve \cite{THUMOS14}. In order to adopt these metrics into unsupervised setting, we use cluster similarity measure(csm)\cite{liao05} which enables us to use any metric in unsupervised setting. It chooses a matching of ground truth labels with predicted labels by searching over matchings and choosing the ones giving highest score. Therefore, $mAP_{csm}$ and $IOU_{csm}$ are our final metrics.

\begin{figure*}[t]
  \includegraphics[width=\textwidth]{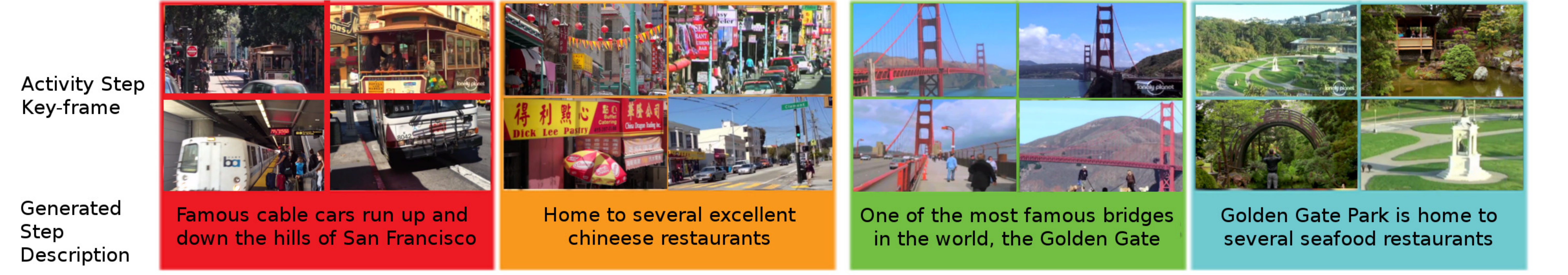}
  \vspace{-6mm}
  \caption{Qualitative results for parsing `Travel San Francisco' category.}
  \vspace{-2mm}
  \label{sf}
\end{figure*}

\vspace{1mm}
\noindent\textbf{Accuracy of the temporal parsing.}
We compute, and plot in Figure\ref{mIOU}, the $IOU_{cms}$ values for all competing algorithms and all categories. We also average over the categories and summarize the results in the Table \ref{averM}. As the Figure~\ref{mIOU} and Table~\ref{averM} suggest, proposed method consistently outperforms the competing algorithms and its variations. One interesting observation is the importance of both modalities reflected in the dramatic difference between the accuracy of our method and its single modal versions.

Moreover, the difference between our method and HMM is also significant. We believe this is due to the ill-posed definition of activities in HMM since the granularity of the activity steps is subjective. In contrast, our method starts with the well-defined definition of finding set of steps which generate the entire collection. Hence, our algorithm do not suffer from granularity problem.
\begin{table}
\caption{Average $IOU_{cms}$ and $mAP_{cms}$ over all categories.}
{\small
\resizebox{\columnwidth}{!}{%
\begin{tabular}{c|cc|cc|cccc}
 & KTS \cite{potapov2014category}    & KTS\cite{potapov2014category}     & HMM     & HMM    & Ours    & Ours     & Ours      & Our  \\
 &  w/ LLF &  w/ Sem &  w/ LLF &  w/Sem &  w/ LLF &  w/o Vis &  w/o Lang &  full \\
 \hline
$IOU_{cms}$  & 16.80 & 28.01      & 30.84 &   37.69   &  33.16 &  36.50 & 29.91& 52.36 \\
$mAP_{cms}$  &  n/a  & n/a        & 9.35  &   32.30   &  11.33 &  30.50 &  19.50 & 44.09 \\
\end{tabular}}}
\normalsize
\label{averM}
\vspace{-5mm}
\end{table}

\vspace{1mm}
\noindent\textbf{Coherency and accuracy of activity step discovery.}
Although $IOU_{cms}$ successfully measures the accuracy of the temporal segmentation, it can not measure the quality of discovered activities. In other words, we also need to evaluate the consistency of the activity steps detected over multiple videos. For this, we use unsupervised version of mean average precision $mAP_{cms}$. We plot the $mAP_{cms}$ values per category in Figure~\ref{mmAP} and their average over categories in Table~\ref{averM}. As the Figure~\ref{mmAP} and the Table~\ref{averM} suggests, our proposed method outperforms all competing algorithms. One interesting observation is the significant difference between semantic and low-level features. Hence, our mid-level features play a key role in linking videos.

\vspace{1mm}
\noindent\textbf{Semantics of activity steps.}
In order to evaluate the role of semantics, we performed a subjective analysis. We concatenated the activity step labels in the grount-truth into a label collection. Then, we ask non-expert users to choose a label for each discovered activity for each algorithm. In other words, we replaced the maximization step with subjective labels. We designed our experiments in a way that each clip received annotations from 5 different users. We randomized the ordering of videos and algorithms during the subjective evaluation. Using the labels provided by subjects, we compute the mean average precision $(mAP_{sem})$.

\begin{table}
\caption{Semantic mean-average-precision $mAP_{sem}$.}
\vspace{-3mm}
{\small
\resizebox{\columnwidth}{!}{%
\begin{tabular}{c|cc|cccc}
            & HMM     & HMM    & Ours    & Ours     & Ours      & Our  \\
            & w/ LLF  &  w/Sem &  w/ LLF &  w/o Vis &  w/o Lang &  full \\ \hline
$mAP_{sem}$ & 6.44   & 24.83  &     7.28 &   28.93  &  14.83    &  39.01 \\
\end{tabular}}}
\normalsize
\vspace{-5mm}
\end{table}

Both $mAP_{cms}$ and $mAP_{sem}$ metrics suggest that our method consistently outperforms the competing ones. There is only one recipe in which our method is outperformed by our baseline of no visual information. This is mostly because of the specific nature of the recipe \emph{How to tie a tie?}. In such videos the notion of object is not useful since all videos use a single object -tie-.

\vspace{1mm}
\noindent\textbf{The importance of each modality.}
As shown in Figure~\ref{mIOU} and \ref{mmAP}, the performance, consistently across all categories, drops when any of the modalities is ignored. Hence, the joint usage is necessary. One interesting observation is the fact that using only language information performed slightly better than using only visual information. We believe this is due to the less intra-class variance in the language modality (\ie, people use same words for same activities). However, it lacks many details(less complete) and is more noisy than visual information. Hence these results validate the complementary nature of language and vision.

\vspace{1mm}
\noindent\textbf{Generalization to generic structured videos}
We experiment the applicability of our method beyond How-To videos by evaluating it on non-How-To categories. In Figure \ref{sf}, we visualize the results for the videos retrieved using the query ``Travel San Francisco". The resulting clusters follow semantically meaningful activities and landmarks and show the applicability of our method beyond How-To queries. It is interesting to note that Chinatown and Clement St ended up in the same cluster; considering the fact that Clement St is known for its Chinese food, this shows successful utilization of semantic connections.
\vspace{-2mm}

%% file: conclusion.tex

\section{Conclusions}
\vspace{-2mm}
In this paper, we tried to capture the underlying structure of human communication by jointly considering visual and language cues. We experimentally validated that given a large-video collection having subtitles, it is possible to discover activities without any supervision over activities or objects. Experimental evaluation also suggested the available noisy and incomplete information is powerful enough to not only discover activities but also describe them. We also think that the resulting discovered knowledge can be effectively used in many domains like multimedia interfaces and robot knowledge bases \cite{robobrain}. 
\vspace{-2mm}
\section{Acknowledgements}
\vspace{-2mm}
We acknowledge the support of ONR award N00014-13-1-0761 and ONR award N000141110389.
\vspace{-2mm}